\documentclass[runningheads]{llncs}
\usepackage{graphicx}
\usepackage{subcaption}
\usepackage{csquotes}
\usepackage[export]{adjustbox}
\usepackage{amsmath, amssymb}
\usepackage{dsfont}
\usepackage{xcolor}
\usepackage{multirow}
\usepackage[linesnumbered, ruled,vlined]{algorithm2e} % own added for pseudo code
\SetKwInOut{KwModel}{Model}
\usepackage{cleveref}

\DeclareMathOperator*{\argmax}{arg\,max}
\begin{document}
\title{Graphhopper: Multi-Hop Scene Graph Reasoning for Visual Question Answering}
\author{Rajat Koner\inst{*1} \and
Hang Li\inst{*1,2} \and
Marcel Hildebrandt \inst{*1,2} 
\and Deepan Das \inst{3} \and
Volker Tresp \inst{1,2} \and
Stephan Günnemann \inst{3}
}
\authorrunning{R. Koner et al.}
\titlerunning{Graphhopper}% Part of RIGHT running header
\institute{Ludwig Maximilian University of Munich, Germany \and
Siemens AG, Germany \and
Technical University of Munich, Germany \\ $\,^*$ equal contribution
}

\maketitle              % typeset the header of the contribution
\begin{abstract}
Visual Question Answering (VQA) is concerned with answering free-form questions about an image. Since it requires a deep semantic and linguistic understanding of the question and the ability to associate it with various objects that are present in the image, it is an ambitious task and requires multi-modal reasoning from both computer vision and natural language processing. We propose Graphhopper, a novel method that approaches the task by integrating  knowledge graph reasoning, computer vision, and natural language processing techniques. Concretely, our method is based on performing context-driven, sequential reasoning based on the scene entities and their semantic and spatial relationships. As a first step, we derive a scene graph that describes the objects in the image, as well as their attributes and their mutual relationships. Subsequently, a reinforcement learning agent is trained to autonomously navigate in a multi-hop manner over the extracted scene graph to generate reasoning paths, which are the basis for deriving answers. We conduct an experimental study on the challenging dataset GQA, based on both manually curated and automatically generated scene graphs. Our results show that we keep up with human performance on manually curated scene graphs. Moreover, we find that Graphhopper outperforms another state-of-the-art scene graph reasoning model on both manually curated  and  automatically generated scene graphs by a significant margin.

\end{abstract}
\section{Introduction}
\label{sec:introduction}
\begin{figure*}[ht]
    \begin{subfigure}{0.52\textwidth}
        \includegraphics[width=\linewidth, height=6.5cm,
        center]{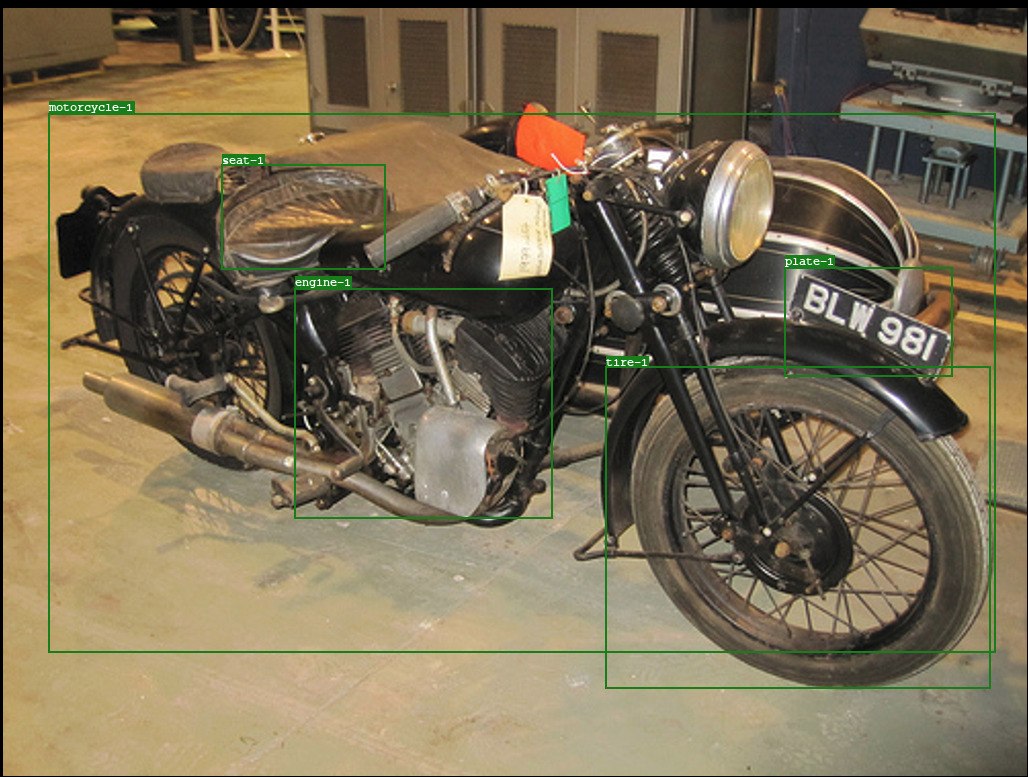} 
        \label{fig:subim1}
    \end{subfigure}
    \begin{subfigure}{0.45\textwidth}
    \vspace{-3mm}
        \includegraphics[width=\linewidth,center]{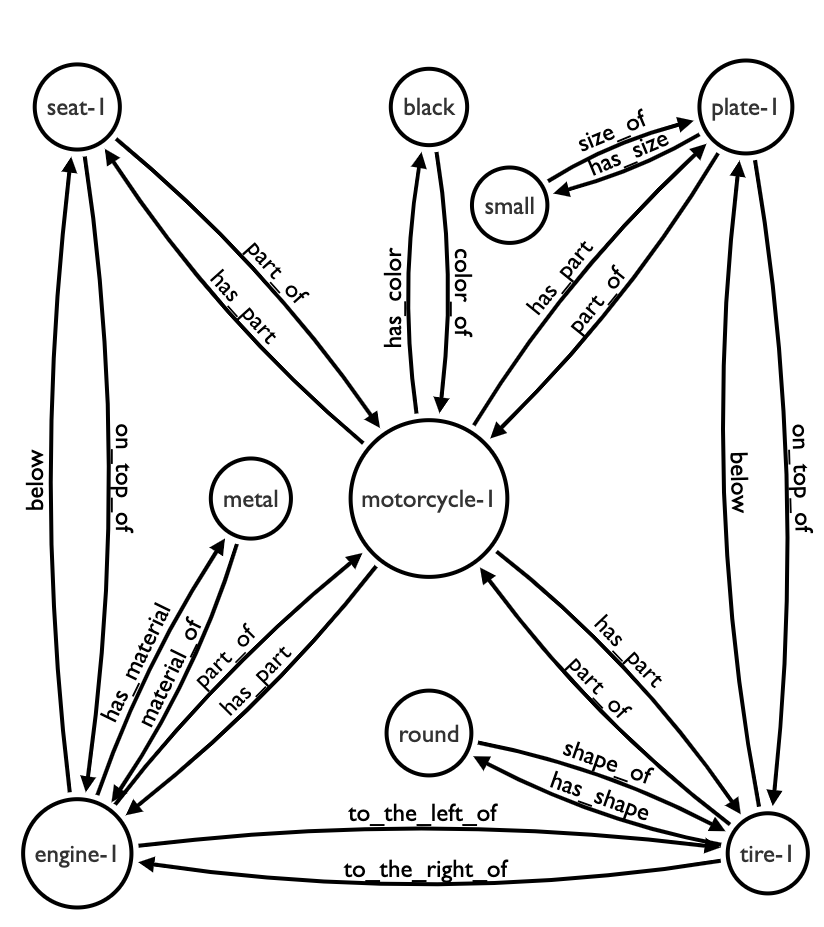}
        \label{fig:subim2}
    \end{subfigure}
    \caption{Example of an image and the corresponding scene graph. Since the scene graph is a directed graph with typed edges, it resembles a knowledge graph and permits the application of  knowledge-base completion techniques.
    %\textsc{I assume the initial hub node is not shown?}
    }
    \label{fig:example_image_sg}
\end{figure*}
Visual Question Answering (VQA) is a challenging task that involves understanding and reasoning over two data modalities, i.e.,  images and natural language. Given an image and a free-form question which formulates a query about the presented scene --- the issue is for the algorithm to find the correct answer.
% to a query which formulates a
% question about the presented scene. %Primarily, a successful VQA should solve three sub task: understanding the image, understanding the free from question and perform reasoning over both data modalities.  

VQA has been studied from the perspective of  scene and knowledge graphs \cite{tang2020unbiased,chen2019counterfactual}, as well as  vision-language reasoning \cite{gan2020large,abbasnejad2020counterfactual}.  To study VQA, various real-world data sets, such as the \textit{VQA} data set \cite{antol2015vqa,krishna2017visual},  have been generated.
It has been argued that, in the \textit{VQA} data set, many of the apparently challenging reasoning tasks can be solved by an algorithm through exploiting trivial prior knowledge,  and thus by shortcuts to proper reasoning
(e.g., clouds are white or doors are made of wood). 
To address these shortcomings,  the \textit{GQA}  dataset \cite{hudson2019gqa} has been developed.
Compared to other real-world datasets, \textit{GQA}  is more suitable for evaluating reasoning abilities since the images and questions are carefully filtered to make the data less prone to biases. 

Plenty of VQA approaches are agnostic towards the explicit relational structure of the objects in the presented scene and rely on monolithic neural network architectures that process regional features of the image separately \cite{anderson2018bottom,yang2016stacked}. While these methods led to promising results on previous datasets, they lack explicit compositional reasoning abilities, which results in weaker performance on more challenging datasets such as \textit{GQA}. 
Other works \cite{teney2017graph,shi2019explainable,hudson2019learning} perform reasoning on explicitly detected objects and interactive semantic and spatial relationships among them. These approaches are closely related to the scene graph representations \cite{johnson2015image} of an image, where detected objects are labeled as nodes and relationships between the objects are labeled as edges.
In this work, we aim to combine VQA techniques with recent research advances in the area of statistical relation learning on knowledge graphs (KGs).
KGs provide human-understandable, structured representations of knowledge about the real world via collections of factual statements. Inspired by multi-hop reasoning methods on KGs such as \cite{minerva,wenhan_emnlp2017,hildebrandt2020reasoning}, we propose Graphhopper, a novel method that models the VQA task as a path-finding problem on scene graphs. The underlying idea can be summarized with the phrase: Learn to walk to the correct answer. More specifically, given an image, we consider a scene graph and train a reinforcement learning agent to conduct a policy-guided random walk on the scene graph until a conclusive inference path is obtained. In contrast to purely embedding-based approaches, our method provides explicit reasoning chains that lead to the derived answers. To sum up, our major contributions are as follows.
\begin{itemize}
    \item Graphhopper is the first VQA method that employs reinforcement learning for multi-hop reasoning on scene graphs.
    \item We conduct a thorough experimental study on the challenging VQA dataset named QGA to show the compositional and \textit{interpretable} nature of our model.
    \item To analyze the reasoning capabilities of our method, we consider manually curated (ground truth) scene graphs. This setting isolates the noise associated with the visual perception task and focuses solely on the language understanding and reasoning task. Thereby, we can show that our method achieves human-like performance.
    \item Based on both the manually curated scene graphs and our own automatically generated scene graphs, we show that Graphhopper outperforms the Neural State Machine (NMS), a state-of-the-art scene graph reasoning model that operates in a setting, similar to Graphhopper.
    % \item To the best of our knowledge, we are the first group to conduct experiments and publish the code on generated scene graphs for the GQA dataset\footnote{Code is available at :\url{https://github.com/rajatkoner08/Graphhopper}}.
\end{itemize}

Moreover, we are the first group to conduct experiments and publish the code on generated scene graphs for the GQA dataset\footnote{Code is available at :\url{https://github.com/rajatkoner08/Graphhopper}}.The remainder of this work is organized as follows. We review related literature in the next section. Section \ref{sec:our_method} introduces the notation and describes the methodology of Graphhopper. Section \ref{sec:experiments} and Section \ref{sec:results} detail an experimental study on the benchmark dataset GQA. Furthermore, through a rigorous study using both manually-curated  ground-truth and generated scene graphs, we examine the reasoning capabilities of  Graphhopper. We  conclude in Section \ref{sec:conclusion}.

\section{Related Work}
\paragraph{Visual Question Answering:} Various models have been proposed that perform VQA on both real-world \cite{antol2015vqa,hudson2019gqa} and artificial datasets \cite{johnson2017clevr}. Currently, leading VQA approaches can be categorized into two different branches: First, monolithic neural networks, which perform implicit reasoning on latent representations obtained from fusing the two data modalities. Second,  multi-hop methods that form explicit symbolic reasoning chains on a structured representation of the data. Monolithic network architectures obtain visual features from the image either in the form of individual detected objects or by processing the whole image directly via convolutional neural networks (CNNs). The derived embeddings are usually scored against a fixed answer set along with the embedding of the question obtained from a sequence model. Moreover, co-attention mechanisms are frequently employed to couple the vision and the language models allowing for interactions between objects from both modalities \cite{kim2018bilinear,anderson2018bottom,cadene2019murel,yu2017multi,zhu2017structured}. Monolithic networks are among the dominant methods on previous real-world VQA datasets such as \cite{antol2015vqa}. 
However, they suffer from the black-box problem and possess limited reasoning capabilities with respect to complex questions that require long reasoning chains (see \cite{chen2019meta} for a detailed discussion).

Explicit reasoning methods combine the sub-symbolic representation learning paradigm with symbolic reasoning approaches over structured representations of the image. Most of the popular explicit reasoning approaches follow the idea of neural module networks (NMNs) \cite{andreas2016neural} which perform a sequence of reasoning steps realized by forward passes through specialized neural networks that each correspond to predefined reasoning subtasks. Thereby, NMNs construct functional programs by dynamically assembling the modules resulting in a question-specific neural network architecture. In contrast to the monolithic neural network architectures described above, these methods contain a natural transparency mechanism via functional programs. However, while NMN-related methods (e.g., \cite{hu2017learning,mao2019neuro}) exhibit good performance on synthetic datasets such as CLEVR \cite{johnson2017clevr}, they require functional module layouts as additional supervision signals to obtain good results. % Hence, NMNs have limited applicability in most real-world settings. In this paper, we propose a novel neuro-symbolic scene graph reasoning approach that leverages representation learning combined with explicit reasoning on scene graphs. 
Closely related to our method is the Neural State Machine (NSM) proposed by \cite{hudson2018compositional}. NSM's underlying idea consists of first constructing a scene graph from an image and treating it as a state machine. Concretely, the nodes correspond to states and edges to transitions. Then, conditioned on the question, a sequence of instructions is derived that indicates how to traverse the scene graph and arrive at the answer. In contrast to NSM, we treat path-finding as a decision problem in a reinforcement learning setting. Concretely, we outline in the next section how extracting predictive paths from scene graphs can be naturally formulated in terms of a goal-oriented random walk induced by a stochastic policy that allows the approach  to balance between exploration and exploitation. Moreover, our framework integrates state-of-the-art techniques from graph representation learning and NLP. This paper only considers basic policy gradient methods, but more sophisticated reinforcement learning techniques will be employed in future works.

\paragraph{Statistical Relational Learning:} Machine learning methods for KG reasoning aim at exploiting statistical regularities in observed connectivity patterns. These methods are studied under the umbrella of statistical relational learning (SRL) \cite{nickel2015review}. In recent years, KG embeddings have become the dominant approach in SRL. The underlying idea is that graph features that explain the connectivity pattern of KGs can be encoded in low-dimensional vector spaces. In the embedding spaces, the interactions among the embeddings for entities and relations can be efficiently modeled to produce scores that predict the validity of a triple. Despite achieving good results in KG reasoning tasks, most embedding-based methods have problems capturing the compositionality expressed by long reasoning chains. This often limits their applicability in complex reasoning tasks. %Multi-hop reasoning methods (also known as path-based methods) follow a different philosophy than embedding-based methods. %The underlying idea is to infer missing knowledge based on sequentially extended inference paths on the KG. 
Recently, multi-hop reasoning methods such as MINERVA  \cite{minerva} and DeepPath \cite{wenhan_emnlp2017} were proposed. Both methods are based on the idea that a reinforcement learning agent is trained to perform a policy-guided random walk until the answer entity to a query is reached. Thereby, the path finding problem of the agent can be modeled in terms of a sequential decision making task framed as a Markov decision process (MDP). The method that we propose  in this work follows a similar philosophy,  in the sense that we train an RL agent to navigate on a scene graph to the correct answer node. However, a conceptual difference is that the agents in MINERVA and DeepPath perform walks on large-scale knowledge graphs exploiting repeating statistical patterns. Thereby, the policies implicitly incorporate approximate rules. In addition, instead of free-form processing questions, the query in the KG reasoning setting is structured as a pair of symbolic entities. That is why we propose a wide range of modifications to adjust our method to the challenging VQA setting.

%\textsc{You don't make the differences very clear}% Scene graph greatly improves the understanding of the images, and facilitates other top label task like action classification\cite{ji2019action}, image generation\cite{johnson2018image} apart from VQA. Over the past years many methods\cite{zellers2018neural,yang2018graph,koner2020relation} has been proposed for scene graph generation by exploring the image context. 

%One such structural prior that underlies our model is that of the probabilistic scene graph [43, 49] which we construct and reason over to answer questions about presented images. Scene graphs provide a succinct representation of the image’s semantics, and have been effectively used for variety of applications such as image retrieval, captioning or generation [43, 54, 45].

\section{Method}
\label{sec:our_method}

The task of VQA is framed as a scene graph traversal problem. Starting from a hub node that is connected to all other nodes, an agent sequentially samples transitions to neighboring nodes on the scene graph until the node corresponding to the answer is reached.  In this way,  by adding transitions  to the current path, the reasoning chain is successively extended. Before describing the decision problem of the agent, we introduce the notation that we use throughout this work.

\paragraph{Notation:}
A scene graph is a directed multigraph where each node corresponds to a scene entity which is either an object associated with a bounding box or an attribute of an object. Each scene entity comes with a type that corresponds to the predicted object or attribute label. Typed edges specify how  scene entities are related to each other. More formally, let $\mathcal{E}$ denote the set of scene entities and consider the set of binary relations $\mathcal{R}$. Then a scene graph $\mathcal{SG} \subset \mathcal{E} \times \mathcal{R} \times \mathcal{E} $ is a collection of ordered triples $(s, p, o)$ – subject, predicate, and object. For example, as shown in Figure \ref{fig:example_image_sg}, the triple \textit{(motorcycle-1, has\_part, tire-1)} indicates that both a motorcycle (subject) and a tire (object) are detected in the image. The predicate \textit{has\_part} indicates the relation between the entities. Moreover, we denote with $p^{-1}$ the inverse relation corresponding to the predicate $p$. For the remainder of this work, we impose completeness with respect to inverse relations in the sense that for every $(s, p, o) \in \mathcal{SG}$ it is implied that $(o, p^{-1}, s) \in \mathcal{SG}$.  %Moreover, we add a so-called hub node (\textit{hub}) to every scene graph which is connected to all other nodes. To answer a binary question, we complement the $\mathcal{SG}$ with two virtual stop nodes - \textit{YES} and
%\textit{NO}. At the end of each trial for a binary question, the agent has to choose from the \textit{YES} or the \textit{NO} node for the last transition.
%\textsc{We need to be care full here: A node in the scene graph is an entity which might have relations of certain types to other entities. We might derive properties of that node: person, tall,  blond and maybe ID; Example: tire is a class, but not an entity; tire1 would be an entity }

\begin{figure}
 \begin{center}
    \includegraphics[width=\textwidth]{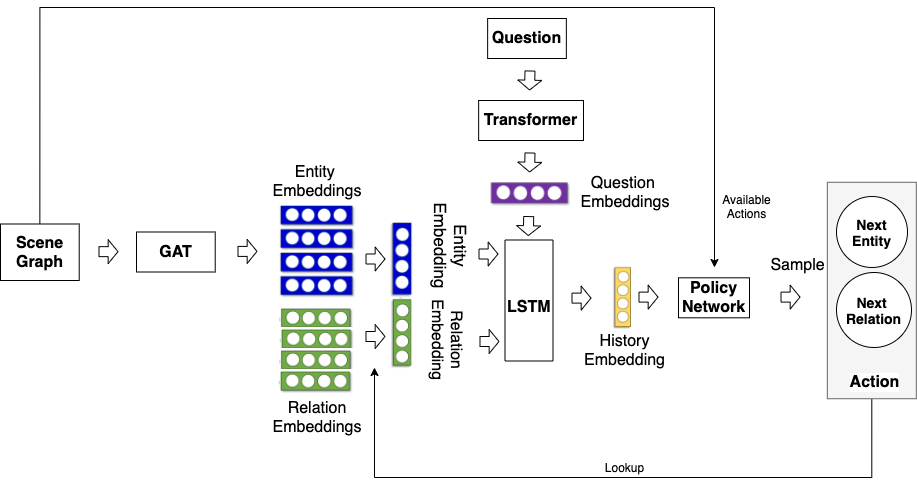}
  \caption{The architecture of our scene graph reasoning module. 
  }
  \label{fig:architecture}   
 \end{center}
\end{figure}

\paragraph{Environment}\sloppy

The state space of the agent $\mathcal{S}$ is given by $\mathcal{E} \times \mathcal{Q}$ where $\mathcal{E}$ are the nodes of a scene graph $\mathcal{SG}$ and $\mathcal{Q}$ denotes the set of all questions. The state at time $t$ is the entity $e_t$ at which the agent is currently located and the question $Q$. Thus, a state $S_t \in \mathcal{S}$ for time $t \in \mathbb{N}$ is represented by $S_t = \left(e_t, Q\right)$. The set of available actions from a state $S_t$ is denoted by $\mathcal{A}_{S_t}$. 
It contains all outgoing edges from the node $e_t$ together with their  corresponding object nodes. 
More formally, $\mathcal{A}_{S_t} = \left\{(r,e) \in \mathcal{R} \times \mathcal{E} :  S_t = \left(e_t, Q\right) \land \left(e_t,r,e\right) \in \mathcal{SG}\right\}\, .$ Moreover, we denote with $A_t \in \mathcal{A}_{S_t}$ the action that the agent performed at time $t$. We include self-loops for each node in $\mathcal{SG}$ that produce  a \textit{NO\_OP}-label. These self-loops  allow the agent to remain at the current location if it reaches the answer node. Furthermore, the  introduction of inverse relations allows agent to transit freely in any direction between two nodes. %To answer binary questions, we also include artificial \textit{yes} and \textit{no} nodes in the scene graph. The agent can transition to these nodes in the final step. Earlier introduction of inverse relations also serve two purposes: First, they allow information to flow in both directions when the representations are formed via the GAT. Second, they allow the agent transition between two connected nodes (i.e., related scene entities) in any direction.
 
The environments evolve deterministically by updating the state according to previous action. Formally, the transition function at time $t$ is given by $\delta_t({S_t},A_t) := \left(e_{t+1}, Q\right)$ with $S_t = \left(e_{t}, Q \right)$ and $A_t = \left(r, e_{t+1}\right)$.

\textbf{Auxiliary Nodes : } In addition to standard entity relation nodes present in a scene graph, we introduce a few auxiliary nodes (e.g. hub node). The underlying rationale for the inclusion  of auxiliary nodes is that they facilitate the walk for the agent or help to frame the QA-task as a goal-oriented walk on the scene graph. These additional nodes are included during run-time graph traversal, but they are ignored during the compile time such as when computing node embedding.
For example, we add a hub node (\textit{hub}) to every scene graph which is connected to all other nodes. The agent then starts the scene graph traversal from a  \textit{hub} with  global connectivity.
Furthermore for a binary question, we add YES and NO nodes to the scene entities that correspond to the final location of the agent. The agent can then transition to either the YES or the NO node. 

\paragraph{Question and Scene Graph Processing} We initialize words in $Q$ with  GloVe embeddings \cite{pennington2014glove} with dimension $d=300$. Similarly we initialize entities and relations  in $\mathcal{SG}$   with the embeddings of  their type labels.
In the scene graph, the node embeddings are passed through a multi-layered graph attention network (GAT) \cite{velivckovic2017graph}. Extending the idea from graph convolutional networks \cite{kipf2016semi} with a  self-attention mechanism, GATs mimic the convolution operator on regular grids where an entity embedding is formed by aggregating node features from its neighbors. Relations and inverse relations between nodes allows context to flow in both ways through GAT. Thus, the resulting embeddings are context-aware, which makes nodes with the same type, but different graph neighborhoods, distinguishable.  To produce an embedding for the question $Q$, we first apply a Transformer \cite{vaswani2017attention},  followed by a mean pooling operation.
%\textsc{For the scene graph: I assume you use the embeddings of the class labels since engine has a representation in Glove but certainly not engine-1; I guess you say labels, which is correct}

Finally, since we added auxiliary \textit{YES} and \textit{NO} nodes to the scene graph for binary questions, we train a feedforward neural network to classify query-type (i.e., questions that query for an object in the depicted scene) and binary questions. This network consists of two fully connected layers with ReLU activation on the intermediate output. We find that it is easy to distingquish between query and binary questions (e.g., query questions usually begin with \textit{What, Which, How}, etc., whereas binary questions usually begin with \textit{Do, Is}, etc.). Since our classifier achieves 99.99\% accuracy we will ignore the error in question classification in the following discussions.

\paragraph{Policy}
We denote the agent's history until time $t$ with the tuple $H_t = \left(H_{t-1}, A_{t-1}\right)$ for $t \geq 1$ and $H_0 = hub$ along with $A_0 = \emptyset$ for $t = 0$. The  history is encoded via a multilayered LSTM \cite{hochreiter1997long}
\begin{equation}
\label{eq:lstm_agent}
        \mathbf{h}_t = \textrm{LSTM}\left(\mathbf{a}_{t-1}\right) \, ,
\end{equation}
where $\mathbf{a}_{t-1} = \left[\mathbf{r}_{t-1},\mathbf{e}_{t}\right] \in \mathbb{R}^{2d}$ corresponds to the embedding of the previous action with $\mathbf{r}_{t-1}$ and $\mathbf{e}_{t}$ denoting the embeddings of the edge and the target node into $\mathbb{R}^{d}$, respectively. 
The history-dependent action distribution is given by
\begin{equation}
\label{eq:policy_agent}
\mathbf{d}_t = \textrm{softmax}\left(\mathbf{A}_t \left(\mathbf{W}_2\textrm{ReLU}\left(\mathbf{W}_1 \left[ \mathbf{h}_t, \mathbf{Q} \right]\right)\right)\right) \, ,
\end{equation}
where the rows of $\mathbf{A}_t \in \mathbb{R}^{\vert \mathcal{A}_{S_t} \vert \times d}$ contain latent representations of all admissible actions. Moreover, $\mathbf{Q} \in \mathbb{R}^{d}$ encodes the question $Q$. The action $A_t = (r,e) \in \mathcal{A}_{S_t}$ is drawn according to $\textrm{categorical}\left(\mathbf{d}_t\right)$. % Equations \eqref{eq:lstm_agent} and \eqref{eq:policy_agent} define a mapping from the history to a distribution over all admissible actions. Thus,  a stochastic policy $\pi_{\theta}$ is induced, where $\theta$ denotes the set of trainable parameters.% in Equations \eqref{eq:lstm_agent} and \eqref{eq:policy_agent}.
Equations \eqref{eq:lstm_agent} and \eqref{eq:policy_agent} induce  a stochastic policy $\pi_{\theta}$, where $\theta$ denotes the set of trainable parameters.% in Equations \eqref{eq:lstm_agent} and \eqref{eq:policy_agent}. 

\paragraph{Rewards and Optimization}
After sampling $T$ transitions, a terminal reward is assigned according to
\begin{equation}
    R = \begin{cases}
 1 &\text{if $e_T$ is the answer to $Q$,} \\
0 &\text{otherwise.}
\end{cases}
\end{equation}
We employ REINFORCE \cite{williams1992simple} to maximize the expected rewards. Thus, the agent's maximization problem is given by
\begin{equation}
\label{eq:objective_agent}
    \argmax_{\theta} \mathbb{E}_{Q \sim \mathcal{T}}\mathbb{E}_{A_1, A_2, \dots, A_N \sim \pi_{\theta}}\left[R \left\vert\vphantom{\frac{1}{1}}\right. e_c \right] \, ,
\end{equation}
where $\mathcal{T}$ denote the set of training questions. During training the first expectation in Equation \eqref{eq:objective_agent} is substituted with the empirical average over the training set. The second expectation is approximated by the empirical average over multiple rollouts.
We also employ a moving average baseline to reduce the variance. Further, we use entropy regularization with parameter $\lambda\in \mathbb{R}_{\geq 0}$ to enforce exploration. During inference, we do not sample paths but perform a beam search with width 20 based on the transition probabilities given by Equation \eqref{eq:policy_agent}.  

Additional details on the model, the training and the inference procedure along with sketches of the algorithms, and a complexity analysis can be found in the supplementary material.

\section{Dataset and Experimental Setup}
\label{sec:experiments}
In this section we introduce the dataset and detail the experimental protocol.

\subsection{Dataset}
\label{subsec:dataset}
The \textit{GQA}  dataset \cite{hudson2019gqa} has been introduced  with the goal of addressing key shortcomings of previous VQA datasets, such as  \textit{CLEVR}  \cite{johnson2017clevr} or the \textit{VQA} dataset \cite{antol2015vqa}.  \textit{GQA} is more suitable for evaluating the reasoning and compositional abilities of a model in a realistic setting. It contains 113K images, and around 1.2M questions split into roughly $80\%/10\%/10\%$ for the training, validation, and testing. The overall vocabulary size consists of 3097 words, including 1702 object classes, 310 relationships, and 610 object attributes. \\
Due to the large number of objects and relationships present in GQA, we used a pruned version of the dataset (see Section~\ref{sec:results}) for our generated scene graph. In this work, we have conducted two primary experiments. First, we report the results on manually curated scene graphs provided in the \textit{GQA} dataset. In this setting, the true reasoning and language understanding capabilities of our model can be analyzed. Afterward, we evaluate the performance of our model with the generated scene graphs on pruned GQA dataset. It shows the performance of our model on noisy generated data. We have used state of the art Relation Transformer Network (RTN) \cite{koner2020relation} for the scene graph generation and DetectoRS \cite{detectors} for object detection. 
We have conducted all the experiments on ``test-dev'' split of the GQA.
\vspace{-3mm}
\paragraph{Question Types: } The questions are designed to evaluate the reasoning abilities such as visual verification, relational reasoning, spatial reasoning, comparison, and logical reasoning. These questions can be categorized either according to structural or semantic criteria. An overview of the different question types is given in supplementary (see Table \ref{tab:experiment_question}).

\subsection{Experimental Setup}

\paragraph{Scene Graph Reasoning: }

Regarding the model parameters, we apply 300 dimensional GloVe embeddings to both the questions and the graphs (i.e., edges and nodes). Moreover, we employ a two-layer GAT \cite{velivckovic2017graph} model. The dropout \cite{srivastava2014dropout} probability of each layer is set to 0.1. The first layer has eight attention heads. Each head has eight latent features which are concatenated to form the output features of that layer. The output layer has eight attention heads with mean aggregation,  so that the output also has 300-dimensional features. We apply dropout with $p=0.1$ to the attention coefficients at each layer. This essentially means that each node is exposed to a stochastically sampled neighborhood during training.
Moreover, we employ a two-layer Transformer \cite{vaswani2017attention} decoder model. The model dimension is set to 300, and the key and query dimensions are both set to 64 with dropout $p=0.1$ . The LSTM of the policy networks consists of a uni-directional layer with hidden size 300. Finally, the agent performs a fixed number of transitions. In question answering, most questions concern one subject to be explored within one reasoning path originated from the start node. Hence, we set the maximum number of steps to 4,  without resetting. By contrast, the binary questions have 8 steps and a reset frequency of 4. In other words, the agent is prompted to the hub node after the fourth step. 

\paragraph{Training the Graphhopper: } In terms of the training procedure, the GAT, the Transformer, and the policy networks are initialized with Glorot \cite{glorot2010understanding} initialization. We train our model with data from the \textit{val\_balanced\_questions} tier. We use a batch size of 64 and sample a batch of questions along with their associated graphs. We collect 20 stochastic rollouts for each question performed in a vectorized form to
utilize parallel computation. For each batch, we collect the rewards when a
complete forward pass is done. Then the gradients are approximated from the
rewards and applied to update the weights.  We employ the Adam optimizer \cite{kingma2014adam} with a
learning rate of $10^{-4}$ for all trainable weights. The coefficient for the action entropy, which balances exploration and exploitation,  starts from
0.2 and decreases exponentially at each step with a factor 0.99. 

Next to other standard Python libraries, we mainly employed PyTorch \cite{paszke2017automatic}. All experiments were conducted on a machine with one NVIDIA RTX 2080 Ti GPU and 64 GB RAM. Training the scene graph reasoner of Graphhopper for 40 epochs on \textit{GQA}  takes around 10 hours, testing about 1 hour. 

\subsection{Performance Metrics}
Along with the accuracy (i.e., Hits@1) on open questions (``Open''), binary questions (yes/no) (``Binary''), and the overall accuracy (``Accuracy''), we also report the additional metric ``Consistency'' (answers should not contradict themselves), ``Validity'' (answers are in the range of a question; e.g., \textit{red} is a valid answer when asked for the color of an object), ``Plausibility'' (answers should be reasonable;  e.g., red is a reasonable color of an apple reasonable, blue is not), as  proposed in \cite{hudson2019gqa}. 

\section{Results and Discussion}
\label{sec:results}

As outlined before, VQA is a challenging task, and there is still a significant performance gap between state-of-the-art VQA methods and human performance on challenging, real-world datasets such as GQA (see \cite{hudson2019gqa}). Similar to other existing methods, our architecture involves multiple components, and it important to be able to analyse the performance of the different modules and processing steps in isolation.  Therefore we first present the results of our experiments on manually curated, ground-truth scene graphs provided in the GQA dataset and compare the performance of Graphhopper against NSM and humans. This setting allows us to isolate the noise from the visual perception component and quantify our methods' reasoning capabilities. Subsequently, we present the results with our own generated scene graphs.

In addition, we also observed that the inclusion of auxiliary nodes helps the agent to achieve efficient performance. \textit{Hub} node performs better compare to starting from any random nodes, as its facilitate easier forward  and backtracking from a node. For binary question instead of YES or NO node, we experimented where the path of the agent was processed by another classifier (e.g., a logistic regression) and the classification logits were assigned as rewards.  However, this led to inferior results; most likely due to the absence of a weight-sharing mechanism and due to the noisy reward signal produced by the classifier. These observations supports our assumption on the role of auxiliary nodes we have used in scene graph.
\paragraph{Reproducing NSM:} \cite{hudson2019learning} proposed the state of the art method named NSM for VQA. NSM is the conceptually most similar method, as it also exploits the scene graph reasoning for VQA. We consider NSM to be  our baseline method for comparison. However, their approach to reasoning is different from ours. To compare the reasoning ability of our method with the same generated scene graph, we tried to reproduce NSM, as the code for NSM is not open-sourced. We have used the the available parameters from \cite{hudson2019learning} and the implementation from \cite{ceyzaguirre4}.\\

%This section we will discuss the performance and reasoning ability of the model based on manually curated and generated scene graph.
\begin{table}
\caption{A comparison of Graphhopper with human performance and NSM based on manually curated scene graphs.
}
\begin{center}
%\resizebox{\textwidth}{!}{
\begin{tabular}{c c c c c c c c}
\hline
Method & Binary & Open & Consistency & Validity & Plausibility & Accuracy \\
\hline
Human \cite{hudson2019gqa}   & 91.2 & 87.4 & \textbf{98.4} & \textbf{98.9} & \textbf{97.2} &  89.3 \\
%TRRNet \cite{yangtrrnet} & 	77.91&	50.22&	89.84&	85.15&	96.47&	63.20 \\
NSM \cite{hudson2019learning} & 	51.03&	18.79&  81.36&	83.69&	79.12&	34.5 \\
Graphhopper   & \textbf{92.18} & \textbf{92.40} & 91.92 & 93.68 & 93.13  & \textbf{92.30} \\
\hline
\end{tabular}
%}
\label{tab:gt_results}
\end{center}
\vspace{-1cm}
\end{table}
\subsection{Results on Manually Curated Scene Graphs}
In this section, we report on an experimental study with Graphhopper on the manually curated scene graphs provided along with the \textit{GQA}  dataset. Table \ref{tab:gt_results} shows the performance of Graphhopper and compares it with the human performance reported in \cite{hudson2019gqa} and with the performance of NSM on the same underlying manually curated scene graphs. We find that Graphhopper strictly outperforms NSM with respect to all performance measures. In particular, on the open questions, the performance gap is significant. Moreover, Graphhopper also slightly outperforms humans with respect to the accuracy on both types of questions.
On the other hand, concerning the supplementary performance measures consistency, validity, and plausibility, Graphhopper is outperformed by humans but nevertheless consistently reaches high values. Overall, these results can be seen as a testament of the reasoning capabilities and establish an upper bound to the performance of Graphhopper. %We expect that improvements in computer vision that lead to more accurate scene graphs will push the performance of Graphhopper towards this upper bound.

%In addition to the model detailed in Section \ref{sec:our_method}, we experimented with various modifications of Graphhopper. Initially, we conducted experiments where the agent's starting location on the scene graph was chosen uniformly at random. Then we ran multiple rollouts with different starting nodes and a joint prediction was formed via a majority vote. However, this procedure is problematic when the scene graph consists of multiple connected components. Therefore, we introduced an artificial hub node as starting location that allows the agent to transition to any other node in the graph. Initially we had concerns towards this approach because the node embedding formed by the GAT are expressive enough so that the agent can learn to make one transition to the answer node and remain there. This would undermine the philosophy of our multi-hop reasoning approach and limit the interpretability. However, we found that the agent still explores the scene graphs in this setting. In many cases, the agent even makes the initial transition to the correct answer node but keeps exploring the neighborhood in order to extract relevant information before returning to the answer node. 

\subsection{Results on automatically generated graph}
The process of generating a graph representation for visual data is a costly and complex procedure. 
%Thus, associated scene graph along with a vision or VQA dataset is rare. Generation of Scene Graph along with object attributes from an image may solve this problem. 
Although the scene graph generation is not the main focus of this work, it constituted one of the major challenges to create good scene graph for GQA due to the following facts:
\begin{itemize}
    \item There is no open source code for GQA scene graph generation or object detection.
    \item A large number of instances and an uneven class distribution in GQA leads to a significant drop in the accuracy compared to existing scene graph  datasets (see \cite{krishna2017visual}).
    \item There is a lack of attribute prediction models in modern object detection frameworks.
\end{itemize}
In this work, we adress all of these challenges as our model's performance is directly dependent on the quality of the scene graph. We will also open-source our code base for transparency and accelerate the development scene graph-based reasoning for VQA. 

\paragraph{Generation of Scene Graph: }
To address these problems, first, we choose two state-of-the-art network, RTN \cite{koner2020relation} for scene graph generation, and DetectoRS \cite{detectors} for object detection. The transformer \cite{vaswani2017attention} based architecture of RTN and its contextual scene graph embedding is most closely related to our architecture and for our future expansion. To make Graphhopper generic to any scene graph generator, we haven't use contextualized embedding from RTN, instead we rely on GAT for contextualization.  
\paragraph{Pruning of GQA: }
GQA has more than 6 times the number of relationships compared to  Visual Genome \cite{krishna2017visual}, which is the most used scene graph generation dataset, and contains
more than 18 times the number of objects compared to the most common object detection dataset COCO \cite{lin2014microsoft}. Also, the class distribution is highly skewed which causes a significant drop in the accuracy for both the object detection and the scene graph generation task. To efficiently prune the number of instances, we take the first 800 classes, 170 relationships, and 200 attributes based on their frequency of occurrence in the training questions and answers. This pruning allows us to reduce more than $60\%$ of the words while covering more than $96\%$ of the combined answers in the training set.

\paragraph{Attribute prediction: }
One of the shortcomings of existing scene graph generation and object detection networks is that they do not predict the attributes (e.g., the color or size of an object) of a detected object. Therefore, we have incorporated the attribute prediction for answering the question on GQA. Contextualized object embedding from RTN \cite{koner2020relation} is used for attribute prediction as
\begin{equation}
    P_{attribute} = \sigma(W(Obj_{context},P_{obj})) \, ,
\end{equation}
where $W$ ,$Obj_{context}$, $P_{obj}$ , $P_{attribute}$ are the weight matrices of a linear layer, the contextual embedding of an object, the probability distribution over all objects and the probability distribution over the attributes. $\sigma$ denotes the sigmoid function.\\
We have trained both the object detector and the scene graph generator on a pruned version of GQA  with their respective default parameters after the prepossessing. This helps to increase the coverage of all the instances (e.g., objects, attributes, relationships ) on training questions from $52\%$ to $77\%$ implying that our generated scene graph now covers $77\%$ of all instances that represent answers to the training questions. 

% \begin{table}[ht!]
% \centering
% \caption{A comparison of our method with generated scene graphs and relations.}
% \begin{tabular}{p{2cm}llllll} 
% \hline
% Method & Binary & Open & Consistency & Validity & Plausibility & \multicolumn{1}{c}{\begin{tabular}[c]{@{}c@{}}Accuracy\\Gen SG ~ ~ ~ ~Overall\end{tabular}}  \\ 
% \hline
% NSM \cite{hudson2019learning}    & ~51.88 &	~19.83 & ~~~~~~0.0 & ~~86.28 & ~~~81.75 & ~~35.34 ~ ~ ~ ~ ~ 35.34                                                                    \\
% Graphhopper  & ~69.48      & ~44.69    & ~~~~83.64           & ~~89.42        & ~~~85.13            & ~~56.69 ~ ~ ~ ~ ~ 48.? \\
% Graphhopper(predicted relations)  & ~85.84      & ~77.27    & ~~~~92.98           & ~~92.26        & ~~~89.50  & ~~81.41 ~ ~ ~ ~ ~ 48.? \\
% \hline
% \end{tabular}
% \label{tab:generated_results}
% \end{table}

\begin{table}
\centering
\caption{A comparison of our method  with NSM,  based on generated scene graphs. Graphhopper (pr) indicates that we employed predicted relations from RTN \cite{koner2020relation}.}
\label{tab:generated_results}
\begin{tabular}{lcccccc} 
\hline
Method             & Binary                 & Open                   & Consistency            & Validity               & Plausibility           & Accuracy                \\ 
\hline
NSM \cite{hudson2019learning}& 51.88                  & 19.83                  & 82.01                  & 86.28                  & 81.75                  & 35.34                   \\
Graphhopper        & \textbf{69.48}                  & \textbf{44.69}                  & \textbf{83.64}                  & \textbf{89.42}                  & \textbf{85.13}                  & \textbf{56.69}                   \\
\hline
Graphhopper (pr) & {85.84} & {77.27} & {92.98} & {92.26} & {89.50} & {81.41}                       \\
\hline
\end{tabular}
\end{table}

\begin{figure}[t]
    \begin{subfigure}{0.315\textwidth}
        \includegraphics[width=\linewidth,center]{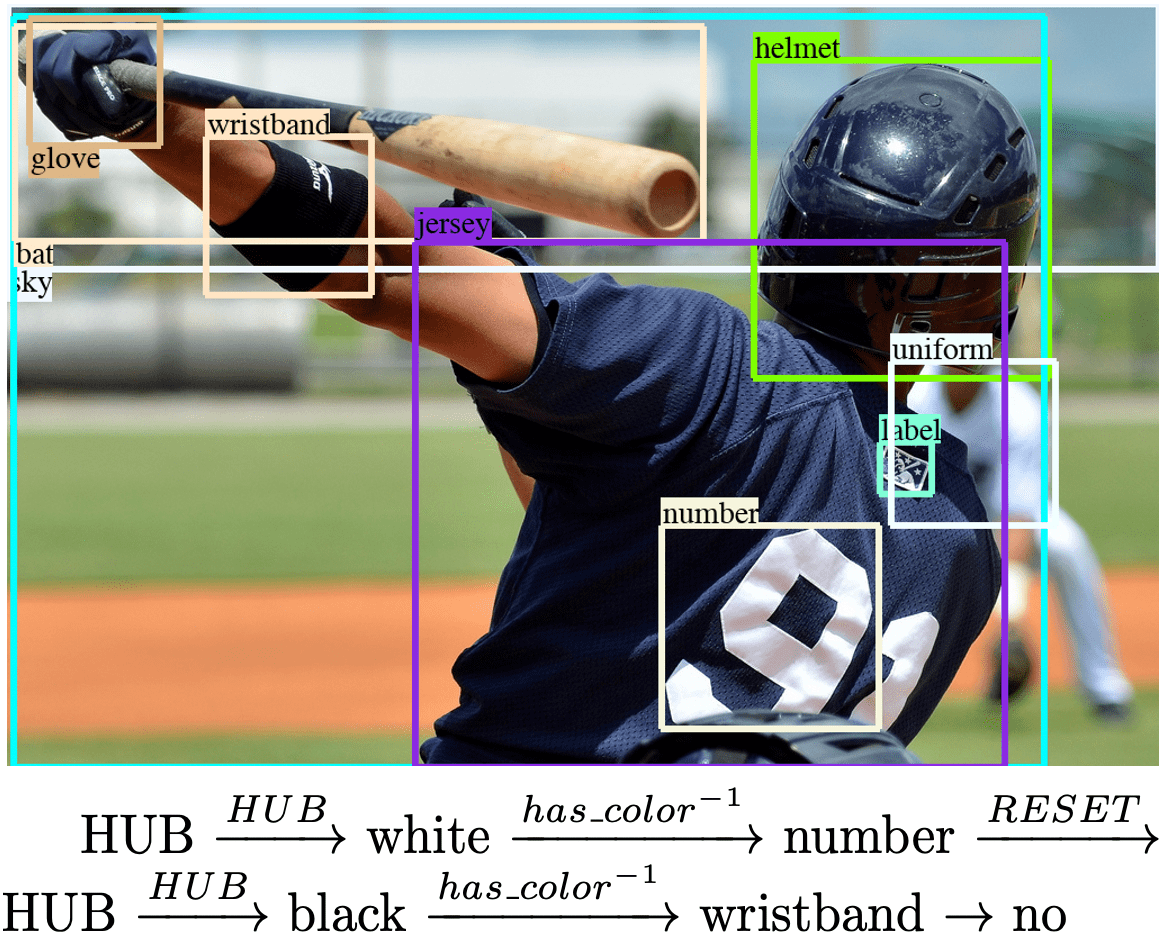} 
        \caption{Question: Is the color of the number the same as that of the wristband? \\Answer: No.}
        \label{subfig:VQA_ex1}
    \end{subfigure}
    \hspace{1mm}
    \begin{subfigure}{0.31\textwidth}
        \includegraphics[width=\linewidth,center]{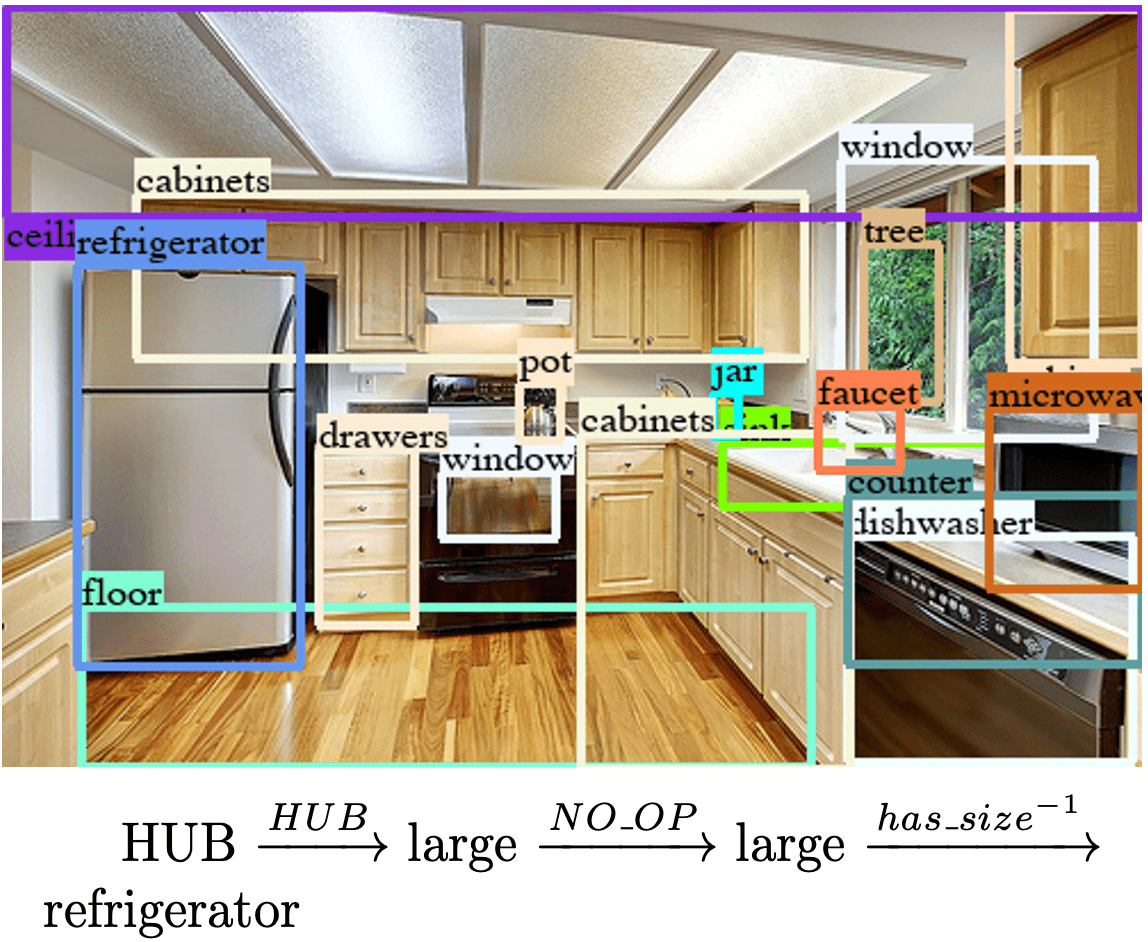}
        \caption{Question: What is the name of the appliance that is not small? \\Answer: Refrigerator.
}
        \label{subfig:VQA_ex2}
    \end{subfigure}
    \hspace{1mm}
    \begin{subfigure}{0.31\textwidth}
        \includegraphics[width=\linewidth,center]{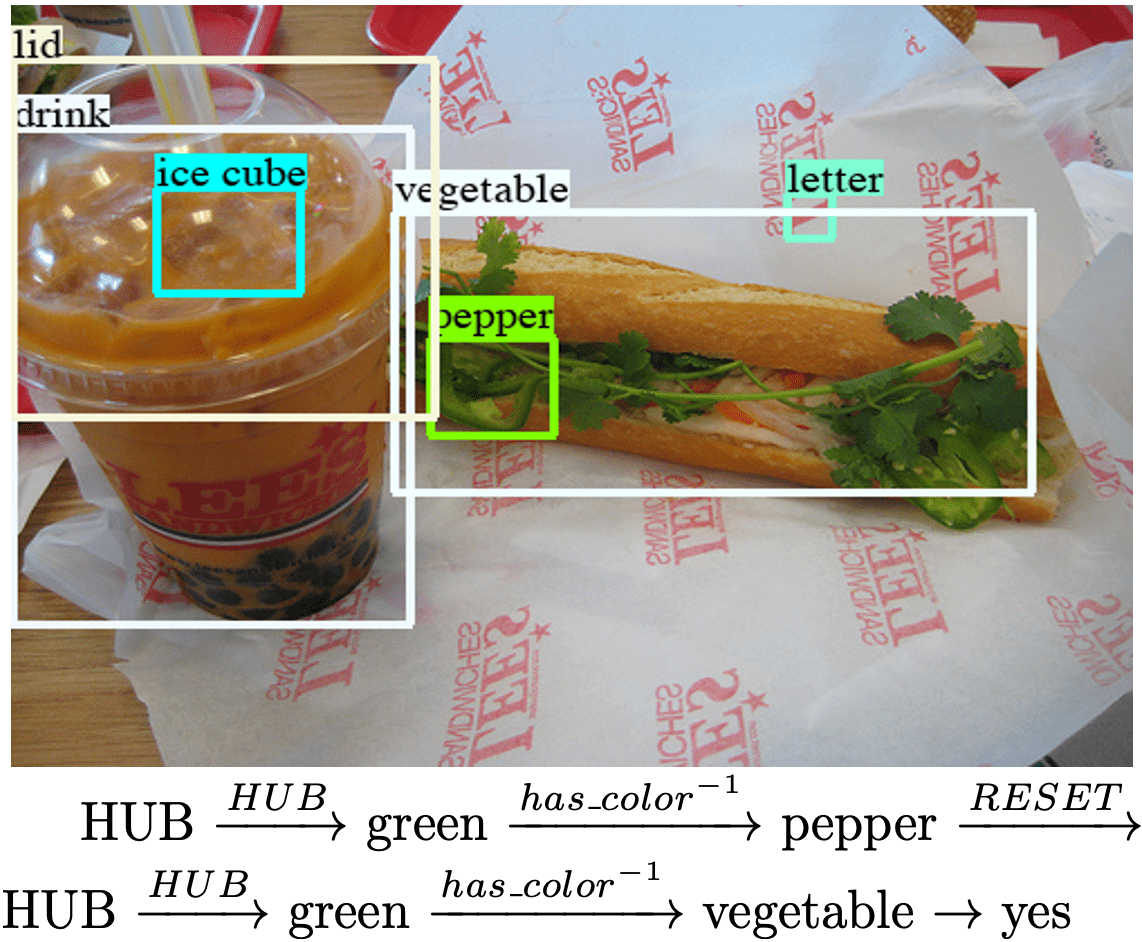}
        \caption{Do both the pepper and the vegetable to the right of the ice cube have green color? \\Answer: Yes.
}
        \label{subfig:VQA_ex3}
    \end{subfigure}
    \caption{Three examples question and the corresponding images and paths.}
    \label{fig:example_paths}
\end{figure}

\begin{figure}[!htbp]
%column number 1 for GT
  \subfloat[Experiments on Manually Curated Scene Graph]{
	\begin{minipage}[c][1.1\width]{
	   0.3\textwidth}
	   \centering
	   \includegraphics[width=1\textwidth]{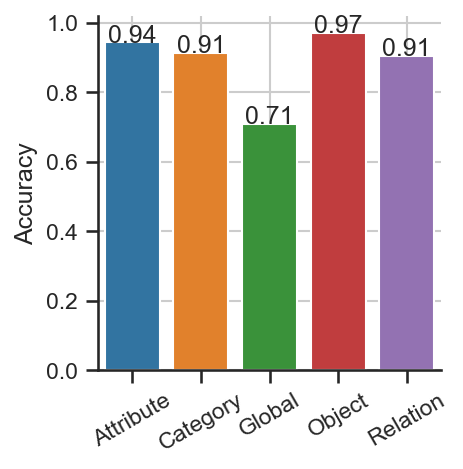}
	\end{minipage}
 \hfill 	
	\begin{minipage}[c][1.1\width]{
	   0.3\textwidth}
	   \centering
	   \includegraphics[width=1\textwidth]{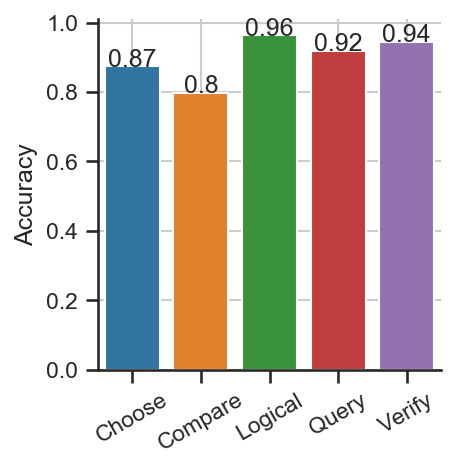}
	\end{minipage}
 \hfill	
	\begin{minipage}[c][1.1\width]{
	   0.3\textwidth}
	   \centering
	   \includegraphics[width=1.1\textwidth]{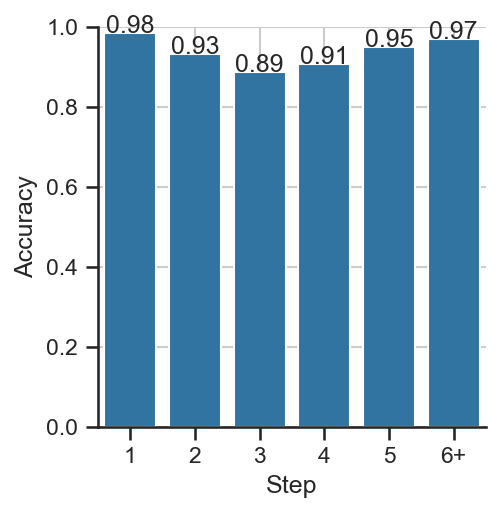}
	\end{minipage}
	\label{fig:gt_reasoning}}\\

    \subfloat[Experiments on Ground Truth objects and predicted relation from RTN \cite{koner2020relation} as Relation predictor. ]{
    	\begin{minipage}[b][1.1\width]{
    	   0.3\textwidth}
    	   \centering
    	   \includegraphics[width=1\textwidth]{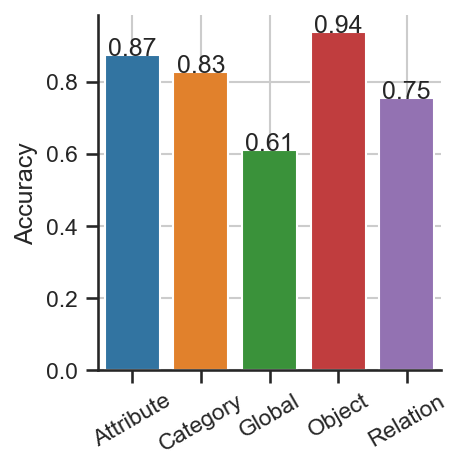}
    	\end{minipage}
    	\begin{minipage}[b][1.1\width]{
    	   0.3\textwidth}
    	   \centering
    	   \includegraphics[width=1\textwidth]{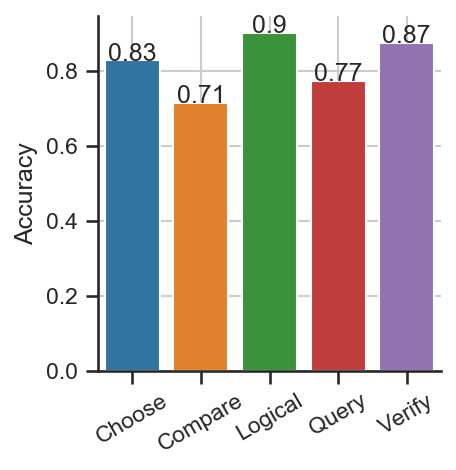}
    	\end{minipage}
    	\begin{minipage}[b][1.1\width]{
    	   0.3\textwidth}
    	   \centering
    	   \includegraphics[width=1.1\textwidth]{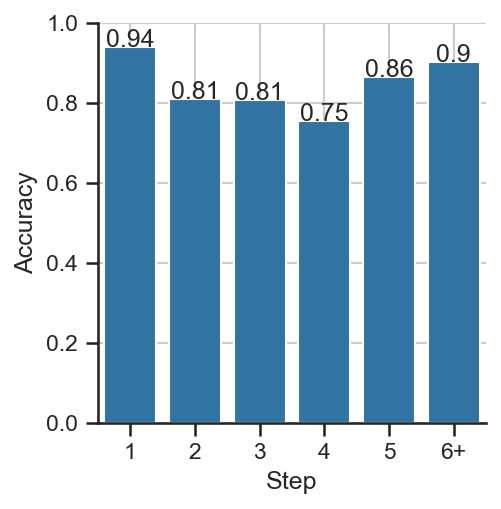}
    	\end{minipage}
    	\label{fig:predr_reasoning}}\\
    	
    \subfloat[Experiments on Generated Scene Graph using DetectoRS \cite{detectors} object detector and RTN \cite{koner2020relation} as Scene Graph generator. ]{
    	\begin{minipage}[b][1\width]{
    	   0.3\textwidth}
    	   \centering
    	   \includegraphics[width=1\textwidth]{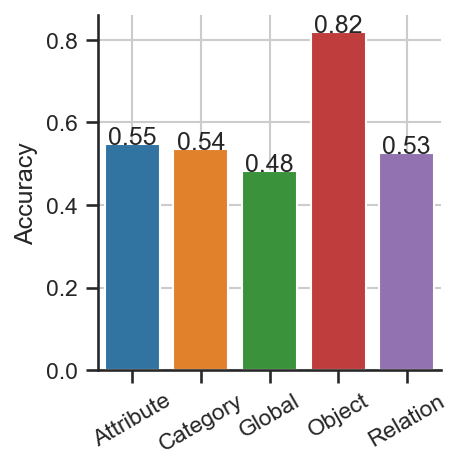}
    	\end{minipage}
    	\begin{minipage}[b][1\width]{
    	   0.3\textwidth}
    	   \centering
    	   \includegraphics[width=1\textwidth]{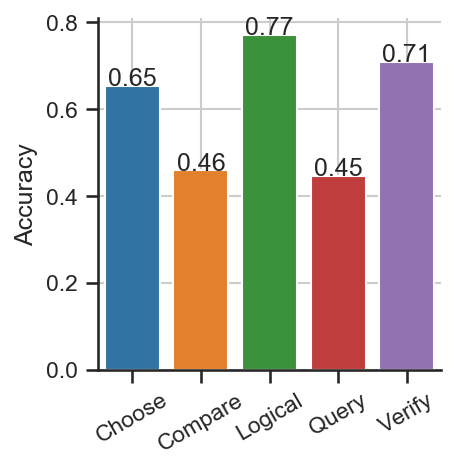}
    	\end{minipage}
    	\begin{minipage}[b][1.1\width]{
    	   0.3\textwidth}
    	   \centering
    	   \includegraphics[width=1.1\textwidth]{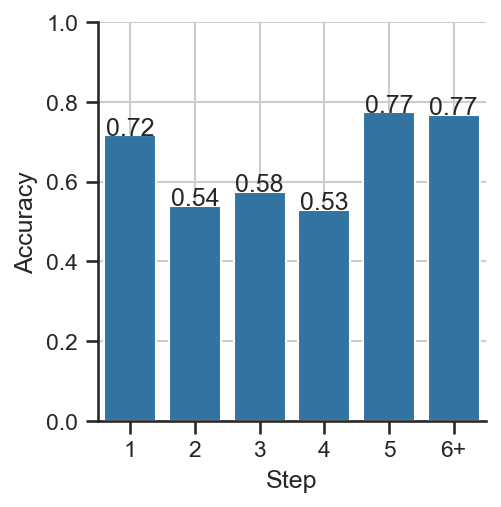}
    	   \label{fig:pred_graph_reasoning}
    	\end{minipage}}
    
\caption{Comparison of the performance of our model on various Scene Graph generation settings, (left) accuracy across various semantic instances (``Attribute'',``Global'',``Relation'' etc) required to answer a question (middle) accuracy on multiple types of question category (``Choose'',``Logical'',``Verify'' etc) and (right) accuracy on minimum number of steps needed to reach the answer node.}
\label{fig:reasoning}
\end{figure}

Table \ref{tab:generated_results}, shows the performance of Graphhopper in two settings: First, with a generated graph where we predict the classes, the attributes, and relationships using our own pipeline. Second, where we only use the predicted relationships from RTN \cite{koner2020relation} (with ground truth objects and attributes). We find that Graphhopper consistently outperforms NSM \cite{hudson2019learning} based on the generated graph. Moreover, in the ``pr'' or predicted relations setting, it achieves an even higher score as the graphs do not contain any misprediction from the object detector. These encouraging results show superior reasoning abilities both on the generated graph and generated relationships between objects. 

\subsection{Discussion on the Reasoning Ability}
\label{sec:reasoning}
To further analyze the reasoning abilities of Graphhopper, Figure \ref{fig:reasoning} disentangles the results according to different types of questions: 5 semantic types (left) and 5 structural types (middle). Moreover, we report the performance of Graphhopper according to the length of the reasoning path (right) (see the supplementary material for additional information). 
Moreover, we show the performance of Graphhopper separately for each of the three scene graph settings that we considered in this work.  Figure \ref{fig:gt_reasoning} shows performance on  a manually curated scene graph that depicts the actual performance in an ideal environment. Figure \ref{fig:predr_reasoning} illustrates the performance based on only the predicted relationships between objects. This setting shows the performance of Graphhopper along with a scene graph generator. Finally, Figure \ref{fig:pred_graph_reasoning} depicts the performance based on the object detector, the scene graph generator, and Graphhopper. First and foremost, we find that Graphhopper consistently achieves high accuracy on all types of questions in every setting. Moreover, we find that the performance of Graphhopper does not suffer if answering the questions requires many reasoning steps. We conjecture that this is because high-complexity questions are harder to answer, but due to proper contextualization of the embeddings (e.g., via the GAT and the Transformer), the agent can extract the specific information that identifies the correct target node. The good performance on these high-complexity questions can be seen as evidence that Graphhopper can efficiently translate the question into a transition on the scene graph hopping until the correct answer is reached.
\paragraph{Examples of Reasoning Path: } Figure \ref{fig:example_paths} shows three examples of scene graph traversals of Graphhopper that lead to the correct answer. One can see in these examples that the sequential reasoning process over explicit scene graph entities makes the reasoning process more comprehensible. In the case of wrong predictions, the extracted path may offer insights into the mechanics of Graphhopper and facilitate debugging.
\section{Conclusion}
\label{sec:conclusion}
We have proposed Graphhopper, a novel method for visual question answering that integrates existing KG reasoning, computer vision, and natural language processing techniques. Concretely, an agent is trained to extract conclusive reasoning paths from scene graphs. %Our experimental findings on the challenging GQA dataset show that MODELNAME outperforms several popular baseline methods while being either more interpretable than existing black-box methods.
To analyze the reasoning abilities of our method, we conducted a rigorous experimental study on both manually curated and generated scene graphs. Based on the manually curated scene graphs we showed that Graphhopper reaches human performance. Moreover, we find that, on our own automatically generated scene graph, Graphhopper outperform another state-of-the-art scene graph reasoning model with respect to all considered performance metrics. In future works, we plan to combine scene graphs with common sense knowledge graphs to further enhance the reasoning abilities of Graphhopper.

%To the best of our knowledge, MODELNAME is the first explicit scene graph reasoning method for VQA based on reinforcement learning. Therefore, we consider MODELNAME to be the first work in a line of research. In this work, we have considered basic policy gradient techniques and we expect that state-of-the-art reinforcement learning methods including extensions such as a cooperative multi-agent systems or sophisticated optimization algorithms will lead to performances gains in the future.
\bibliographystyle{splncs04}
\bibliography{bibliography}

\begin{thebibliography}{10}
\providecommand{\url}[1]{\texttt{#1}}
\providecommand{\urlprefix}{URL }
\providecommand{\doi}[1]{https://doi.org/#1}

\bibitem{abbasnejad2020counterfactual}
Abbasnejad, E., Teney, D., Parvaneh, A., Shi, J., Hengel, A.v.d.:
  Counterfactual vision and language learning. In: Proceedings of the IEEE/CVF
  Conference on Computer Vision and Pattern Recognition. pp. 10044--10054
  (2020)

\bibitem{anderson2018bottom}
Anderson, P., He, X., Buehler, C., Teney, D., Johnson, M., Gould, S., Zhang,
  L.: Bottom-up and top-down attention for image captioning and visual question
  answering. In: Proceedings of the IEEE conference on computer vision and
  pattern recognition. pp. 6077--6086 (2018)

\bibitem{andreas2016neural}
Andreas, J., Rohrbach, M., Darrell, T., Klein, D.: Neural module networks. In:
  Proceedings of the IEEE Conference on Computer Vision and Pattern
  Recognition. pp. 39--48 (2016)

\bibitem{antol2015vqa}
Antol, S., Agrawal, A., Lu, J., Mitchell, M., Batra, D., Lawrence~Zitnick, C.,
  Parikh, D.: Vqa: Visual question answering. In: Proceedings of the IEEE
  international conference on computer vision. pp. 2425--2433 (2015)

\bibitem{cadene2019murel}
Cadene, R., Ben-Younes, H., Cord, M., Thome, N.: Murel: Multimodal relational
  reasoning for visual question answering. In: Proceedings of the IEEE
  Conference on Computer Vision and Pattern Recognition. pp. 1989--1998 (2019)

\bibitem{chen2019counterfactual}
Chen, L., Zhang, H., Xiao, J., He, X., Pu, S., Chang, S.F.: Counterfactual
  critic multi-agent training for scene graph generation. In: Proceedings of
  the IEEE/CVF International Conference on Computer Vision. pp. 4613--4623
  (2019)

\bibitem{chen2019meta}
Chen, W., Gan, Z., Li, L., Cheng, Y., Wang, W., Liu, J.: Meta module network
  for compositional visual reasoning. arXiv preprint arXiv:1910.03230  (2019)

\bibitem{minerva}
Das, R., Dhuliawala, S., Zaheer, M., Vilnis, L., Durugkar, I., Krishnamurthy,
  A., Smola, A., McCallum, A.: Go for a walk and arrive at the answer:
  Reasoning over paths in knowledge bases using reinforcement learning. In:
  ICLR (2018)

\bibitem{ceyzaguirre4}
Eyzaguirre, C.: Nsm. \url{https://github.com/charlespwd/project-title} (2019)

\bibitem{gan2020large}
Gan, Z., Chen, Y.C., Li, L., Zhu, C., Cheng, Y., Liu, J.: Large-scale
  adversarial training for vision-and-language representation learning. arXiv
  preprint arXiv:2006.06195  (2020)

\bibitem{glorot2010understanding}
Glorot, X., Bengio, Y.: Understanding the difficulty of training deep
  feedforward neural networks. In: Proceedings of the thirteenth international
  conference on artificial intelligence and statistics. pp. 249--256 (2010)

\bibitem{hildebrandt2020reasoning}
Hildebrandt, M., Serna, J.A.Q., Ma, Y., Ringsquandl, M., Joblin, M., Tresp, V.:
  Reasoning on knowledge graphs with debate dynamics. arXiv preprint
  arXiv:2001.00461  (2020)

\bibitem{hochreiter1997long}
Hochreiter, S., Schmidhuber, J.: Long short-term memory. Neural computation
  \textbf{9}(8),  1735--1780 (1997)

\bibitem{hu2017learning}
Hu, R., Andreas, J., Rohrbach, M., Darrell, T., Saenko, K.: Learning to reason:
  End-to-end module networks for visual question answering. In: Proceedings of
  the IEEE International Conference on Computer Vision. pp. 804--813 (2017)

\bibitem{hudson2019learning}
Hudson, D., Manning, C.D.: Learning by abstraction: The neural state machine.
  In: Advances in Neural Information Processing Systems. pp. 5901--5914 (2019)

\bibitem{hudson2018compositional}
Hudson, D.A., Manning, C.D.: Compositional attention networks for machine
  reasoning. arXiv preprint arXiv:1803.03067  (2018)

\bibitem{hudson2019gqa}
Hudson, D.A., Manning, C.D.: Gqa: A new dataset for real-world visual reasoning
  and compositional question answering. arXiv preprint arXiv:1902.09506  (2019)

\bibitem{johnson2017clevr}
Johnson, J., Hariharan, B., van~der Maaten, L., Fei-Fei, L., Lawrence~Zitnick,
  C., Girshick, R.: Clevr: A diagnostic dataset for compositional language and
  elementary visual reasoning. In: Proceedings of the IEEE Conference on
  Computer Vision and Pattern Recognition. pp. 2901--2910 (2017)

\bibitem{johnson2015image}
Johnson, J., Krishna, R., Stark, M., Li, L.J., Shamma, D., Bernstein, M.,
  Fei-Fei, L.: Image retrieval using scene graphs. In: Proceedings of the IEEE
  conference on computer vision and pattern recognition. pp. 3668--3678 (2015)

\bibitem{kim2018bilinear}
Kim, J.H., Jun, J., Zhang, B.T.: Bilinear attention networks. In: Advances in
  Neural Information Processing Systems. pp. 1564--1574 (2018)

\bibitem{kingma2014adam}
Kingma, D.P., Ba, J.: Adam: A method for stochastic optimization. arXiv
  preprint arXiv:1412.6980  (2014)

\bibitem{kipf2016semi}
Kipf, T.N., Welling, M.: Semi-supervised classification with graph
  convolutional networks. arXiv preprint arXiv:1609.02907  (2016)

\bibitem{koner2020relation}
Koner, R., Sinhamahapatra, P., Tresp, V.: Relation transformer network. arXiv
  preprint arXiv:2004.06193  (2020)

\bibitem{krishna2017visual}
Krishna, R., Zhu, Y., Groth, O., Johnson, J., Hata, K., Kravitz, J., Chen, S.,
  Kalantidis, Y., Li, L.J., Shamma, D.A., et~al.: Visual genome: Connecting
  language and vision using crowdsourced dense image annotations. International
  Journal of Computer Vision  \textbf{123}(1),  32--73 (2017)

\bibitem{lin2014microsoft}
Lin, T.Y., Maire, M., Belongie, S., Hays, J., Perona, P., Ramanan, D.,
  Doll{\'a}r, P., Zitnick, C.L.: Microsoft coco: Common objects in context. In:
  European conference on computer vision. pp. 740--755. Springer (2014)

\bibitem{mao2019neuro}
Mao, J., Gan, C., Kohli, P., Tenenbaum, J.B., Wu, J.: The neuro-symbolic
  concept learner: Interpreting scenes, words, and sentences from natural
  supervision. arXiv preprint arXiv:1904.12584  (2019)

\bibitem{nickel2015review}
Nickel, M., Murphy, K., Tresp, V., Gabrilovich, E.: A review of relational
  machine learning for knowledge graphs. Proceedings of the IEEE
  \textbf{104}(1),  11--33 (2015)

\bibitem{paszke2017automatic}
Paszke, A., Gross, S., Chintala, S., Chanan, G., Yang, E., DeVito, Z., Lin, Z.,
  Desmaison, A., Antiga, L., Lerer, A.: Automatic differentiation in pytorch
  (2017)

\bibitem{pennington2014glove}
Pennington, J., Socher, R., Manning, C.D.: Glove: Global vectors for word
  representation. In: Empirical Methods in Natural Language Processing (EMNLP).
  pp. 1532--1543 (2014), \url{http://www.aclweb.org/anthology/D14-1162}

\bibitem{detectors}
Qiao, S., Chen, L.C., Yuille, A.: Detectors: Detecting objects with recursive
  feature pyramid and switchable atrous convolution. arXiv preprint
  arXiv:2006.02334  (2020)

\bibitem{shi2019explainable}
Shi, J., Zhang, H., Li, J.: Explainable and explicit visual reasoning over
  scene graphs. In: Proceedings of the IEEE Conference on Computer Vision and
  Pattern Recognition. pp. 8376--8384 (2019)

\bibitem{srivastava2014dropout}
Srivastava, N., Hinton, G., Krizhevsky, A., Sutskever, I., Salakhutdinov, R.:
  Dropout: a simple way to prevent neural networks from overfitting. The
  journal of machine learning research  \textbf{15}(1),  1929--1958 (2014)

\bibitem{tang2020unbiased}
Tang, K., Niu, Y., Huang, J., Shi, J., Zhang, H.: Unbiased scene graph
  generation from biased training. In: Proceedings of the IEEE/CVF Conference
  on Computer Vision and Pattern Recognition. pp. 3716--3725 (2020)

\bibitem{teney2017graph}
Teney, D., Liu, L., van Den~Hengel, A.: Graph-structured representations for
  visual question answering. In: Proceedings of the IEEE Conference on Computer
  Vision and Pattern Recognition. pp.~1--9 (2017)

\bibitem{vaswani2017attention}
Vaswani, A., Shazeer, N., Parmar, N., Uszkoreit, J., Jones, L., Gomez, A.N.,
  Kaiser, {\L}., Polosukhin, I.: Attention is all you need. In: Advances in
  neural information processing systems. pp. 5998--6008 (2017)

\bibitem{velivckovic2017graph}
Veli{\v{c}}kovi{\'c}, P., Cucurull, G., Casanova, A., Romero, A., Lio, P.,
  Bengio, Y.: Graph attention networks. arXiv preprint arXiv:1710.10903  (2017)

\bibitem{williams1992simple}
Williams, R.J.: Simple statistical gradient-following algorithms for
  connectionist reinforcement learning. Machine learning  \textbf{8}(3-4),
  229--256 (1992)

\bibitem{wenhan_emnlp2017}
Xiong, W., Hoang, T., Wang, W.Y.: Deeppath: A reinforcement learning method for
  knowledge graph reasoning. In: Proceedings of the 2017 Conference on
  Empirical Methods in Natural Language Processing (EMNLP 2017). ACL,
  Copenhagen, Denmark (9 2017)

\bibitem{yang2016stacked}
Yang, Z., He, X., Gao, J., Deng, L., Smola, A.: Stacked attention networks for
  image question answering. In: Proceedings of the IEEE conference on computer
  vision and pattern recognition. pp. 21--29 (2016)

\bibitem{yu2017multi}
Yu, Z., Yu, J., Fan, J., Tao, D.: Multi-modal factorized bilinear pooling with
  co-attention learning for visual question answering. In: Proceedings of the
  IEEE international conference on computer vision. pp. 1821--1830 (2017)

\bibitem{zhu2017structured}
Zhu, C., Zhao, Y., Huang, S., Tu, K., Ma, Y.: Structured attentions for visual
  question answering. In: Proceedings of the IEEE International Conference on
  Computer Vision. pp. 1291--1300 (2017)

\end{thebibliography}

\newpage
\appendix

\section{Details on Model Training and Inference}

\subsection{Training Details}\label{sec:method_train}
In order to optimize the training objective given by Equation \eqref{eq:objective_agent}, we use REINFORCE \cite{williams1992simple} to obtain the gradient approximation
\begin{equation}
    \hat{g} = \sum_{t=0}^{T-1}{\nabla_\theta \log \pi_\theta(A_t|S_t) 
    (\sum_{t^{'}=t}^{T-1}{R_{t^{'}}\gamma^{t^{'}-t} - B(S_t)})}\, ,
\end{equation}
where $\gamma$ is the discount factor for the reward. The gradient of the
weights are aggregated over multiple rollouts. To reduce the variance, we adopt a moving average baseline function
$B(S_t)$. The baseline function is an approximation of the value of a state $S_t$. We could have employed more sophisticated methods such as advantage network or actor-critic algorithm. However, we find the current baseline works sufficiently well. Formally, the baseline function consists of a non-trainable
variable $b$ and a hyperparameter $\lambda$. The baseline is updated by $b_{t+1}
= \lambda b_{t} + (1-\lambda) r_t$ at each optimization step. Another technique
that affects the training speed is the reward normalization. Concretely, the
accumulated rewards at each time step for each rollout are collected and
normalized after subtraction of the baseline value.

We introduce a regularization term on the entropy of the resulting probability
distribution from the policy network $\pi_{\theta}(A_t|S_t)$, which
enforces that the agent explores the SG. The regularization is controlled by a
hyperparameter $\beta$. In addition, we apply exponential decay to $\beta$ during
training so that $\beta$ converges to zero. 

Moreover, we use the chain rule to calculate the gradients of the parameters of the graph encoder
(GAT) $\theta_{GAT}$ and the question encoder (Transformer)
$\theta_{Transformer}$. The weight updates can be performed via gradient ascent,
$\theta \leftarrow \theta + \eta \hat{g}$ or more advanced optimization methods
such as Adam \cite{kingma2014adam}.

\begin{algorithm}[ht!]
    \SetAlgoLined
    \KwIn{Question $Q$, Scene Graph $\mathcal{SG} \subset \mathcal{E} \times \mathcal{R} \times \mathcal{E}$}
    \KwModel{ Policy Network with $\theta := \{\theta_{GAT}, \theta_{Transformer}, \theta_{Agent} \}$, 
    Baseline with $b$}
    \BlankLine
    \For(\tcp*[h]{Loop over epochs}){$i \leftarrow 0$ \KwTo $N$}{
        Initialize $Q$ and $\mathcal{SG}$ with GloVe embeddings\;
        $Q \leftarrow \mbox{GAT}(Q)$ \tcp{Update the question with the question encoder}
        $\mathcal{SG} \leftarrow \mbox{Transformer}(\mathcal{SG})$ \tcp{Update the SG with the graph encoder}
        $C \leftarrow []$ \tcp{Initialize the trajectory buffer}
        \For(\tcp*[h]{Loop over samples}){$r \leftarrow 0$ \KwTo $N$}{
            $\tau \leftarrow []$ \tcp{Initialize the trajectory}
            $E_0 \leftarrow$ hub \tcp{Initialize the start position}
            $A_0 \leftarrow$ dummy \tcp{Initialize the dummy start action}
            \For(\tcp*[h]{Loop over time steps}){$t \leftarrow 0$ \KwTo $T$}{
                \If(\tcp*[h]{Restart and prompt the agent to the hub node}){$t\% \Delta ==0$}{
                    \tcp{so that the agent is aware of its own action}
                    $E_{t+1} \leftarrow$ hub \tcp{Set next nodes to the hub node}
                    $A_{t+1} \leftarrow$ dummy   \tcp{Set next actions to the dummy return action}
                }
                Sample an action ($A_t$, $E_{t+1}$) from $d_t$ 
                $\tau$.append($A_t$, $E_t$) \tcp{Extend the trajectory}
                $E_{t} \leftarrow E_{t+1}$ \tcp{Move the agent to the next entity}
            }
            $C$.append($\tau$) \tcp{Collect the trajectory}
        }
        $r \leftarrow R(C)$ \tcp{Gather rewards}
        $g \leftarrow \sum_{t=0}^{T-1}{\nabla_\theta \log \pi_\theta(a_t|s_t) 
        (\sum_{t^{'}=t}^{T-1}{r_{t^{'}}\gamma^{t^{'}-t} - b(s_t)})}$ \tcp{Approximate gradients}
        $\theta \leftarrow \theta + \eta g$ \tcp{Update the policy network}
        $b \leftarrow b + (1-\lambda) r$ \tcp{Update the baseline function}
    }
        \caption{Training regime}
        \label{tab:train}
\end{algorithm}

\subsection{Inference}
Beam search is used to infer the answer to a given question. Our inference approach is based on evaluating how likely specific
paths are appearing among all  possible paths with a fixed length.
More specifically, given an input question, the agent's initial location is given by the hub
node. At each time step, the agent scores the next permissible actions based on the
learned policy. The value of action represents the transition probability
from the current node to a target node. Next, we keep the top $k$ (also known as
beam width) paths among all possible transitions and move the agent to the
corresponding targets. This computation is iteratively performed until the
maximum number of transitions is reached. In the end, we obtain multiple
rollouts ranked by the path probabilities. The target node (i.e. the last node)
of the path is regarded as an answer candidate. Unlike Monte Carlo sampling
which does not consider path probabilities, beam search yields better answer
candidates, as it always chooses the best choice within the search region.
The algorithm for inference is summarized in table \ref{tab:inference}.

\paragraph{Inference Complexity}
The inference of our method is computationally efficient. Unlike other methods
that need to iterate through each candidate answer for a final prediction, we
only need to run the inference once so that the score of each answer is
obtained. Let $d$ denote the embedding dimension of the words and entities.
Analytically, the embedding stage has asymptotic complexity $\mathcal{O}(|\mathcal{E}| +
|\mathcal{R}| + |Q|)$. For the GAT, the implementation of a single attention head
 and multi-head attention is similar. In particular, they have the same time
complexity $\mathcal{O}(|\mathcal{E}|dd^{'} + |\mathcal{R}|d^{'})$. The
computation of the question encoding is given by $\mathcal{O}(|\mathcal{Q}|^2d)$. It
is efficient as it only runs once for each question and is used for arbitrary times
during random walks. Also, the length of the questions $Q$ is usually short (less than 30 words).
Finally, during the random walk sampling, the agents complxity is given by $\mathcal{O}(T(D^2 +
|\mathcal{E}|d))$, where $d$ is dominant. The inference time depends largely on the path length. 

\begin{algorithm}[ht!]
    \SetAlgoLined
    \KwIn{Question $Q$, Scene graph $\mathcal{SG} \subset \mathcal{E} \times \mathcal{R} \times \mathcal{E}$}
    \KwOut{Answer}
    \BlankLine
    Initialize $Q$ and $\mathcal{SG}$ with GloVe embeddings\;
    $Q \leftarrow \mbox{GAT}(Q)$ \tcp{Update the question with the question encoder}
    $\mathcal{SG} \leftarrow \mbox{Transformer}(\mathcal{SG})$ \tcp{Update the SG with the graph encoder.}
    $P \leftarrow []$ \tcp{Initialize the probability register}
    $\tau \leftarrow []$ \tcp{Initialize the trajectory}
    $E_0 \leftarrow$ hub \tcp{Initialize the start position}
    $A_0 \leftarrow$ dummy \tcp{Initialize the dummy start action}
    \For(\tcp*[h]{Loop over time steps}){$t \leftarrow 0$ \KwTo $T$}{
        \For(\tcp*[h]{Loop over rollouts}){$r \leftarrow 0$ \KwTo $N$}{
            \If(\tcp*[h]{Restart and prompt the agent to the hub node}){$t\% \Delta ==0$}{
                \tcp{so that the agent is aware of its own action}
                $E_{t+1} \leftarrow$ hub \tcp{Set next nodes to the hub node}
                $A_{t+1} \leftarrow$ dummy \tcp{Set next actions to the dummy return action}
            }
            Forward pass through the policy network to generate candidate actions $\{(A_t, E_t)\}$ 
            along with their probabilities $\{p_i\}$
            $\tau$.append($\{(A_t, E_t)\}$) \tcp{Extend the trajectory}
            $P$.append($\{p_i\}$) \tcp{Store corresponding probabilities}
        }
        $indicies \leftarrow argmax(P, k)$ \tcp{Filter indices of top k probabilities from P} 
        $\tau \leftarrow \tau[indices]$ \tcp{Choose top k paths ranked by their probabilities}
        $E_{t+1} \leftarrow e \in \tau$ \tcp{Conduct corresponding transitions}
    }
    Prediction $\leftarrow \tau[0]$ \tcp{Predict the end entity of the top path as the answer}
    \caption{Inference with beam search}
    \label{tab:inference}
\end{algorithm}

\subsection{Complexity Analysis}
For analyzing the complexity of our method, we provide all the parameters contained in
the building blocks. Moreover, we present the number of operations of a forward
pass - the complete run that derives the answer from a given $Q$ and
$\mathcal{SG}$. They are listed in the table \ref{tab:operations}.

\begin{table}[htb]
    \begin{tabular}{l|l|l|l}
        \hline
        Group & Name & No. Parameters & No. Operations\\
        \hline
        Word Embeddings* & Entity & $N_e\times d$ & $\mathcal{O}(N)$ \\
         & Relation & $N_r\times D$ & $\mathcal{O}(N)$ \\
        \hline
        GAT & Conv layer weight & $d \times H_1$ & $\mathcal{O}(BHN_e)$\\
         & Conv layer attention & $d \times H_1$ & $\mathcal{O}(BHN_e)$\\
         & Conv layer bias & $H_1$ & $\mathcal{O}(BHN_e)$\\
        \hline
        Transformer & Positional encoder &$d\times d$ & $\mathcal{O}(N_e)$\\
         & Layer self attention $(qkv)$ &$H_t(512)\times d$ & $3\times H_t \times d$\\
         & Self attn norm $(W,b)$ &$d$ & $2\times d$\\
         & Layer enc attn &$H_t(512)\times d$ & $3\times H_t \times d$\\
         & Enc attn norm $(W,b)$ &$d$ & $2\times d$\\
         & Pos ffn 1 $(W,b)$ & $d\times d + d$&$d \times d + d$ \\
         &Pos ffn 2 $(W,b)$ & $d\times d + d$&$d \times d + d$ \\
         &Pos ffn norm$(W,b)$ & $2d$ & $2\times d$ \\
         & Enc attn norm $(W,b)$ &$d$ & $2\times d$\\
        \hline
        Agent-MLP & Dense 0 & $4d \times 4d + 4d$ & $(H\times 4d + 4d)\times T$ \\
         & Dense 1 & $2d \times 4d + 2d$ & $(H\times 2d + 2d)\times T$ \\
        \hline
        Agent-LSTM & Lstm\_cell $W_{ih}$ & $4d\times 16d$ & $(4d\times 16d) \times T$\\
         & Lstm\_cell $_{ih}$ & $16d$ & $(16d) \times T$\\
         & Lstm\_cell $W_{hh}$ & $4d\times 16d$ & $(4d\times 16d) \times T$\\
         & Lstm\_cell $b_{hh}$ & $16d$ & $(16d) \times T$\\
        \hline
    \end{tabular}
    \caption[Overview of model parameters and operations]{An overview of the number parameters and the asymptotic number of operations for the individual modules. The batch size is indicated by $B$.  $T$ corresponds to the number of time steps. $D$ and $H$ denote the embedding size and
    hidden size, respectively. Blocks are marked with a "*" if their weights are not trainable.}
    \label{tab:operations}
\end{table}

\section{Additional Details on the Dataset GQA}
In this section we describe various question category and their type. We list the question based on semantic and structural categories. We further grouped them based on their entity type like object, attribute, category etc. Table \ref{tab:experiment_question}, describes the detailed list of question category.
\begin{table}[htb]
  \caption[List of question examples in the GQA dataset]{List of question examples in the GQA dataset.}\label{tab:experiment_question}
    \centering
    \begin{tabular}{|l|l|l|p{2.5in}|}
        \hline
        Category & Type & Description & Example\\ \hline
        \multirow{5}{*}{Semantics} & Object & Existence of object & Are there any doors that are not made of metal?\\ \cline{2-4}
        & Attribute & Property about an object & Does the soap dispenser that is to the right of the other soap dispenser have small size and white color?\\ \cline{2-4}
        & Category & Identify an object class & What kind of animal is standing? \\ \cline{2-4}
        & Relation & Relationship of object& What is the food that is to the left of the white object that is to the left of the chocolate called? \\ \cline{2-4}
        & Global & Overall scene property & Which place is it? \\ \hline
        \multirow{5}{*}{Structural} &Query& Open-form question& What type of furniture is to the left of the silver device which is to the left of the helmet? \\ \cline{2-4}
        & Choose & Choose from alternatives &  What are the floating people in the ocean doing, riding or swimming?\\ \cline{2-4}
        & Verify & Simple yes/no question & Are there statues above the brass clock that is on the building? \\ \cline{2-4}
        & Compare & Comparison of objects & Are the drawers made of the same material as the cages? \\ \cline{2-4}
        & Logical & And/or operators & Are both the giraffe near the building and the giraffe that is to the left of the tray standing? \\ \hline
    \end{tabular}
\end{table}

\end{document}